\documentclass[letterpaper, 10 pt, conference]{ieeeconf}  

\IEEEoverridecommandlockouts                              

\pdfoutput=1

\usepackage{graphicx}
\usepackage{multicol}
\usepackage{multirow}
\usepackage{amsmath}
\usepackage{amsfonts}
\usepackage{siunitx}
\usepackage{subcaption}
\usepackage{float}
\usepackage{hyperref}
\usepackage{wrapfig,lipsum,booktabs}
\usepackage[dvipsnames]{xcolor}
\graphicspath{ {./teaser/}}

\title{\LARGE \bf
LiDAR guided Small obstacle Segmentation 
}

\author{Aasheesh Singh$^{*1}$, Aditya Kamireddypalli$^{*1}$, Vineet Gandhi$^{1}$ and K Madhava Krishna$^{1}$ 
\thanks{* Equal Contribution}
\thanks{$^{1}$ KCIS, International Institute of Information Technology Hyderabad, India 
        {\tt\small aasheeshdtu@gmail.com}}%
}
\begin{document}

\maketitle
\thispagestyle{empty}
\pagestyle{empty}


\begin{abstract}
Detecting small obstacles on the road is critical for autonomous driving. In this paper, we present a method to reliably detect such obstacles through a multi-modal framework of sparse LiDAR(VLP-16) and Monocular vision. LiDAR is employed to provide additional context in the form of confidence maps to monocular segmentation networks. We show significant performance gains when the context is fed as an additional input to monocular semantic segmentation frameworks. We further present a new semantic segmentation dataset to the community, comprising of over 3000 image frames with corresponding LiDAR observations. The images come with pixel-wise annotations of three classes off-road, road, and small obstacle. We stress that precise calibration between LiDAR and camera is crucial for this task and thus propose a novel Hausdorff distance based calibration refinement method over extrinsic parameters. As a first benchmark over this dataset, we report our results with 73 \% instance detection up to a distance of 50 meters on challenging scenarios. Qualitatively by showcasing accurate segmentation of obstacles less than 15 cms at 50m depth and quantitatively through favourable comparisons vis a vis prior art, we vindicate the method's efficacy. Our project and dataset is hosted at \href{https://small-obstacle-dataset.github.io/}{\emph{https://small-obstacle-dataset.github.io/}}


\end{abstract}
\section{INTRODUCTION}
\label{Intro}
Within the realm of perception for autonomous cars, small obstacle detection is a critical problem. Small obstacles fall precariously on the border of being classified as drivable space or obstacles. It is prudent for the planning module of an autonomous car to be informed of the small obstacles in its environment.

A straightforward extension of semantic segmentation methods that label semantically contiguous regions accurately, while occupying a tiny part of the overall image space has proven difficult \cite{guo2018small}. Since such small area classes occupy an insignificant portion in the predicted label space, they accrue meager error costs that result in diminished gradient magnitudes, providing negligible updates to the model parameters. To overcome these, methods such as \cite{eigen2015predicting, badrinarayanan2017segnet} propose cost terms sensitive to small area class labels (median frequency balancing), which enjoy limited success. However, such methods have not showcased their results on obstacles lying on the road but small obstacle classes like poles and traffic signs seen from afar. Obstacles lying on the road pose a greater challenge because such instances are rare in carefully curated autonomous driving datasets. Additionally, they may not be easy to spot if the appearance is similar to the road (an example is shown in Figure~\ref{fig:teaser}, where large stones lie on a cemented road). In this paper, we develop an alternate paradigm - one that fuses LiDAR and Image inputs effectively for on-road small obstacle detection.

In our method, LiDAR scans, available from the sparse Velodyne Puck 16, are used to detect possible obstacle regions in a scan through a discontinuity detector. A contiguous set of 3D points between two discontinuities form a candidate region where a small obstacle may reside. These candidate regions are projected onto the image plane through LiDAR-Camera extrinsics and are further blurred through the Gaussian blur operator. Such blurred regions in the image are assigned probability scores that represent the confidence of a small obstacle and henceforth called as confidence maps. The confidence map along with the input RGB image forms the four-channel input to a deep convolutional network architecture. The network - supervised on three classes, namely the road class, off-road class, and small obstacle class, - can accurately detect small obstacle regions hitherto difficult for a purely image-based classification framework (a motivating example is shown in Figure~\ref{fig:teaser}). We show significant improvement in detection and classification accuracy of small obstacle regions by such fusion of LiDAR and image-based sensing modalities vis a vis image only classifier architectures. Formally our paper makes the following contributions:

\begin{figure}[t]
   \centering
\begin{tabular}{cc}
\includegraphics[width=0.45\linewidth]{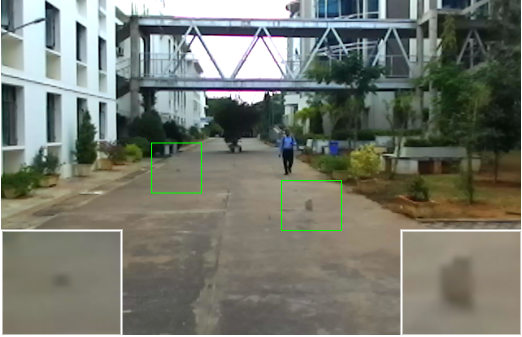}&
\includegraphics[width=0.45\linewidth]{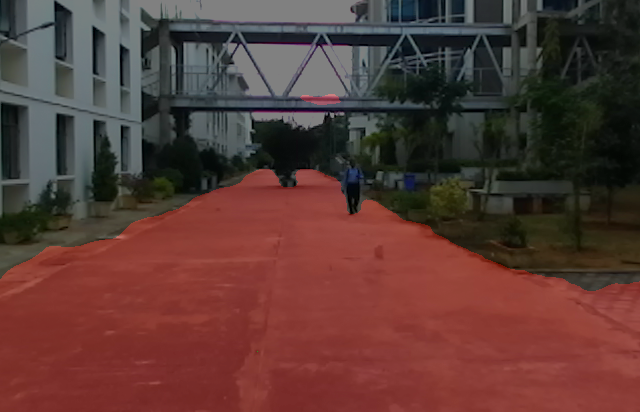}\\ 
 {\small (a)} & {\small (b)} \\
\includegraphics[width=0.45\linewidth]{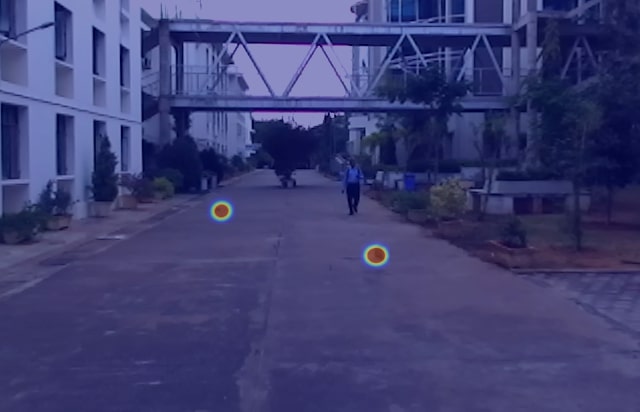}&
\includegraphics[width=0.45\linewidth]{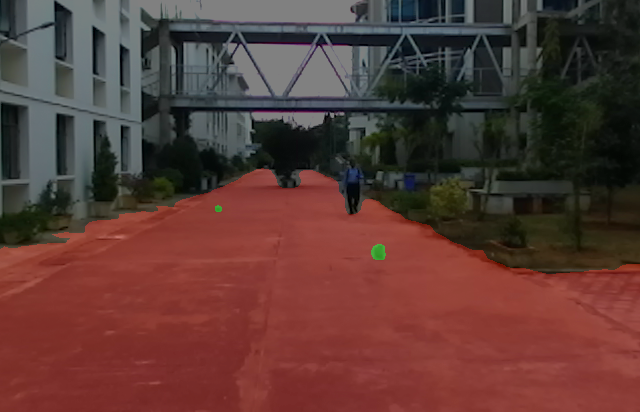}\\
 {\small  (c)} & {\small (d)}
\end{tabular}
    \caption{Example detections: (a) Original image with two small obstacles (green rectangles). (b) Monocular RGB baseline fails to detect both the obstacles. (c) The context (confidence map) generated using LiDAR point cloud. (d) The proposed method (combining monocular image with LiDAR context) successfully detects both the obstacles. }
\label{fig:teaser} 
\end{figure}

\begin{figure*}[t]
\centering
\includegraphics[width=0.99\textwidth]{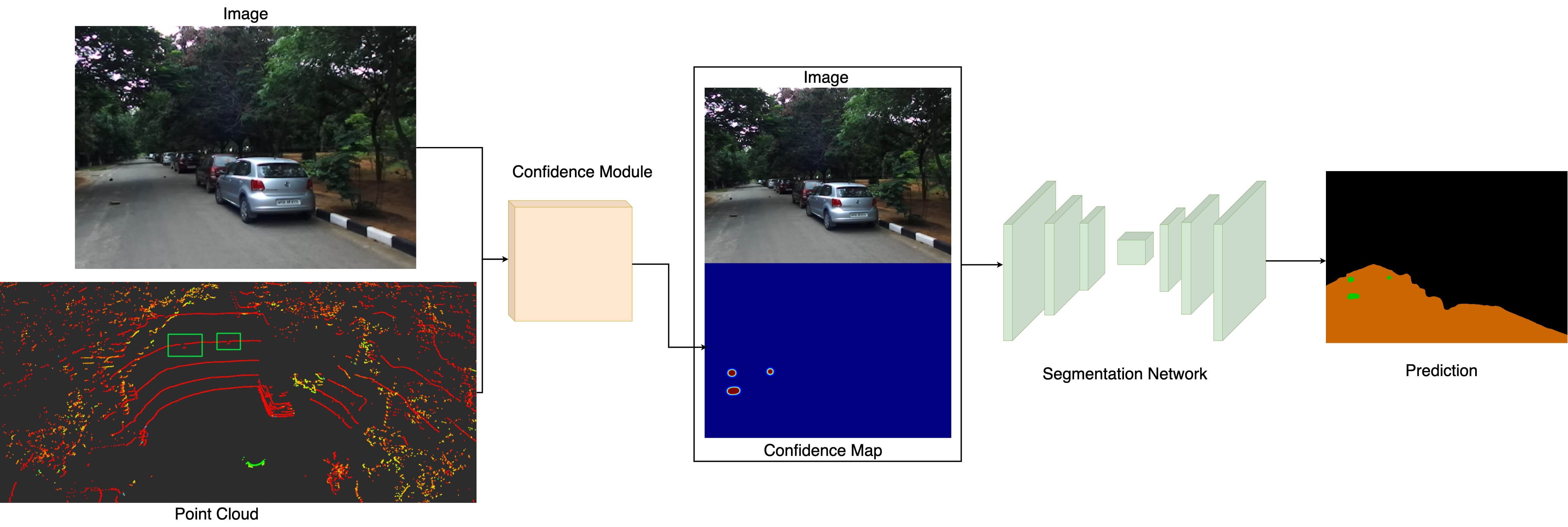}
\caption{The proposed small obstacle detection pipeline. The confidence module generates a confidence map based on the point cloud. The result is concatenated with the current image and used to predict a segmentation mask.}
\label{fig:pipeline}
\end{figure*}

\begin{enumerate}
    \item A novel LiDAR-Camera small obstacle segmentation dataset that comprises over 3000 frames with per pixel annotation of 3 classes, on-road, off-road, and small obstacle along with LiDAR-Camera calibration extrinsics. This dataset can also be viewed as an image only or LiDAR only dataset for small obstacle segmentation.
    \item A novel pipeline that combines confidence map information obtained from LiDAR point clouds along with RGB channel inputs to obtain significant boost over purely image-based architectures such as \cite{DeepLab, HRNet}. It is important to note that sparse LiDAR scans can miss small obstacles entirely, and hence it is not prudent to rely on the current LiDAR input alone. This is overcome by propagating prior LiDAR confidence maps to the current frame, such that continuous per frame small obstacle detection is deemed possible even when LiDAR scans miss the small obstacle completely. We portray these results in section \ref{sec:Expt}. The use of the sparse Puck 16 LiDAR over its denser 64 scan-line counterpart, which is more amenable to point cloud segmentation, is yet another cornerstone of this work.
    \item Detailed ablation studies across alternative methods for various modules that constitute the overall pipeline, such as LiDAR-Camera calibration, confidence map generation, and temporal propagation of confidence maps.
    \item A novel Haussdorf distance-based markerless LiDAR-Camera calibration refinement that significantly improves training time and accuracy as a result of improved extrinsics.
    \item And finally, to the best of our knowledge, this is the first such method that combines sparse LiDAR data with a monocular image to segment small obstacles, unlike most segmentation approaches that operate exclusively on only image or dense point cloud data.
    
\end{enumerate}


\section{RELATED WORK}
\label{Related}
The field of semantic segmentation has seen tremendous success over the past few years. The success can be attributed to the applicability of CNN based encoder-decoder architectures~\cite{badrinarayanan2017segnet, DeepLab, HRNet} and the availability of large datasets which allow efficient training~\cite{cityscapes, everingham2015pascal}. However, handling class imbalance has always remained a challenge as small objects usually contribute less to the segmentation loss. Many semantic segmentation works follow a relatively simple cost-sensitive approach via an inverse frequency rebalancing scheme, e.g.,~\cite{xu2014tell} or median frequency re-weighting~\cite{eigen2015predicting}. The small classes, such as pole and signboards, in autonomous driving datasets like Cityscapes are unable to supervise a network to detect small-obstacles lyingon the road. Apart from class imbalance (and class mismatch), monocular small obstacle detection also faces the challenge of weaker appearance cues (Figure ~\ref{fig:teaser}). To highlight these concerns, we use two state of the art image segmentation networks DeepLab~\cite{DeepLab}and HRNet~\cite{HRNet} as our baselines. Both networks were pretrained using the Cityscape dataset and were fine-tuned on the proposed IIIT Small obstacle detection dataset. 

Specific efforts have been made in the area of on-road small obstacle detection. The carefully curated Lost and Found dataset~\cite{pinggera2016lost} has been pivotal for the progress in this direction. Most of the approaches augment the appearance cues with a depth map obtained using a stereo camera. The work by Pinggera \emph{et al.}~\cite{pinggera2016lost} performs statistical hypothesis tests in disparity space directly on stereo image data, assessing free space and obstacle hypotheses on independent local patches. Ramos \emph{et al.}~\cite{ramos2017detecting} combines deep learning approaches with geometric cues. MergeNet~\cite{mergenet} proposes a novel architecture combining local structural cues obtained from a Stripe based network with a full image output. MergeNet takes inspiration from Stixel based processing~\cite{pfeiffer2011towards} and demonstrates the ability to be trained from only a few images (as low as 135). The major limitation of these approaches is the reliance on inaccurate depth maps, especially in low texture scenarios or in cases where the obstacle appearance is similar to that of the road. 

In this paper, we employ a LiDAR + RGB pipeline for on-road small obstacle detection. LiDAR data has been successfully used for problems like road object segmentation~\cite{wu2018squeezeseg}, curb detection~\cite{zhao2012curb} etc. Combining RGB data has shown to bring improvements in both the tasks of semantic segmentation~\cite{meyer2019sensor} and object detection~\cite{qi2018frustum} (in contrast to pure LiDAR methods). The fusion of the two modalities particularly shows improvements for detecting objects at a long-range (\textgreater40m-70m) over their RGB and LiDAR only counterparts~\cite{meyer2019sensor}. Our work is in a similar spirit. Furthermore, the small obstacle detection problem is difficult to be solved using LiDAR alone because (a) the small obstacle may not intersect with LiDAR rings, especially when using sparse LiDAR (as used in our work) and (b) even if the discontinuities are detected the overall extent of the object will be unknown in the image space. We propose a method fusing sparse 16 channel LiDAR with monocular data and provide pixel-level segmentation in the image space. 


Our work is also related to the literature in LiDAR and RGB calibration~\cite{mirzaei20123d,dhall2017lidar}. We show that minor calibration errors can be extremely problematic for the task of small obstacle detection. Consequently, we propose a method for fine refinement of extrinsic parameters for LiDAR-RGB calibration. The intuition of the approach is to align the detected LiDAR discontinuities on the road with nearby ground truth small obstacle segmentation labels in image space. The refinement algorithm leverages Hausdorff distance for the task and is agnostic to the original calibration algorithm.

\section{METHODOLOGY}\label{Method}
\subsection{Overview}
\begin{figure}[t]
\centering
\includegraphics[width=0.35\textwidth]{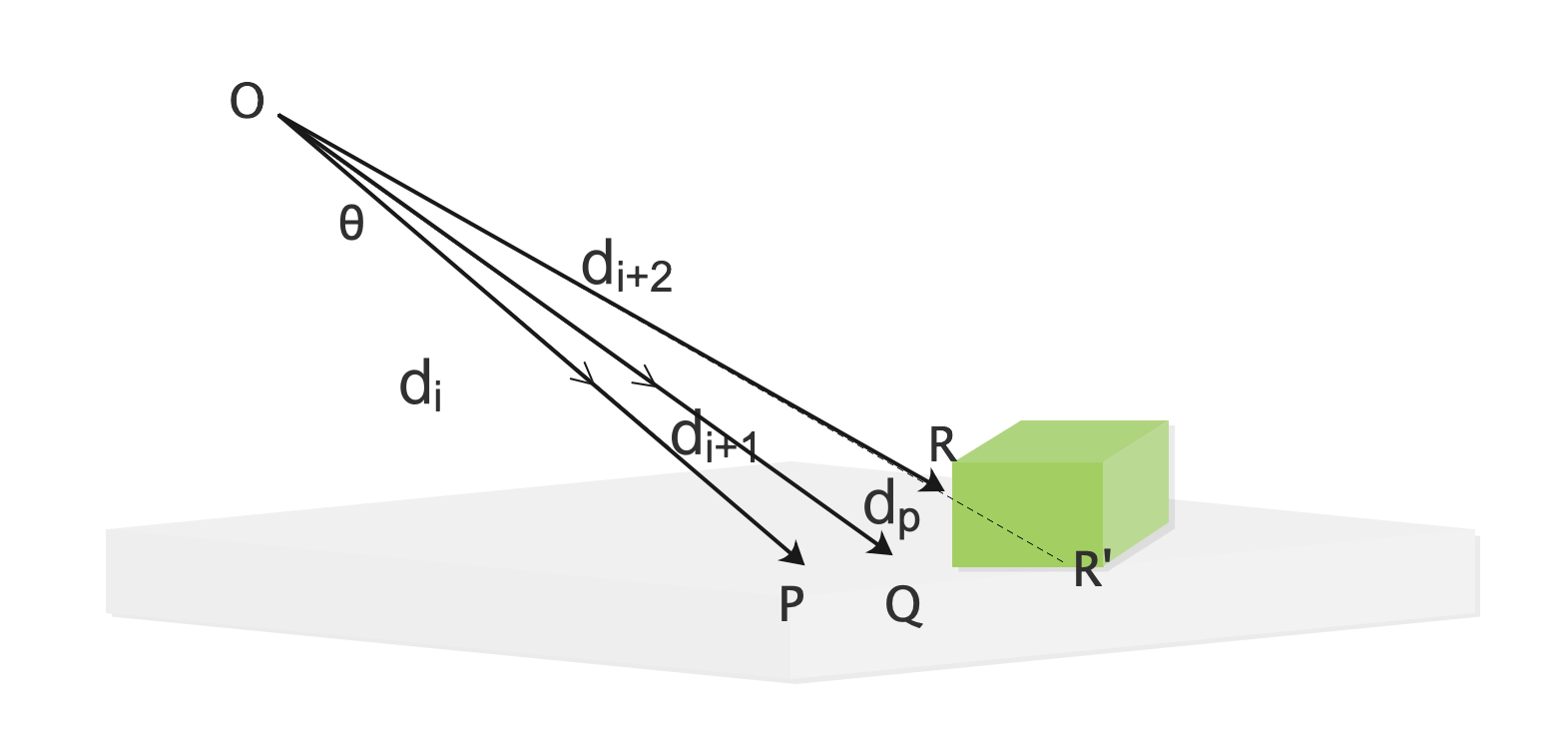}
\caption{Break point identification (for point $R$). The points $P$, $Q$ and $R$ are consecutive points that lie on the same LiDAR ring. $O$ represents the centre of the LiDAR sensor. $OP$, $OQ$, $OR$ and $OR^\prime$ corresponds to $d_{i}$, $d_{i+1}$, $d_{i+2}$ and $d_{p}$ respectively.}
\label{fig:breakpoint}
\end{figure}
The proposed pipeline is illustrated in Figure~\ref{fig:pipeline}. At a given time, the input is an RGB image and the corresponding LiDAR point cloud. The output is a three-class (road, small obstacle, and off-road) semantic segmentation map. At first, the LiDAR point cloud is used to detect discontinuities (breakpoints) on the road using geometric reasoning (the detected discontinuities are illustrated with green rectangles in Figure~\ref{fig:pipeline}). The detected discontinuities are then projected onto the image space and are used to generate a pixel level confidence map. Since the LiDAR used in our experiments is sparse, it may miss out on some of the small obstacles (in cases where no LiDAR point falls on them). The confidence map is further augmented using temporal propagation of discontinuities detected in previous frames. The confidence map with the RGB image (4D input) and the corresponding ground truth maps are then used for training a deep convolutional network for semantic segmentation. We now describe each of these steps in detail.


\subsection{Obstacle Confidence Maps from LiDAR}
\label{sec:geometric}
\textbf{Breakpoint Detection:} We identify geometric break points within each ring of the Point Cloud, where a ring is defined as one complete 360$^\circ$ scan for a given channel of the LiDAR. Our sensor VLP-16 gives 16 such rings within a vertical resolution of 30$^\circ$ (-15 to 15). We use the approach discussed in \cite{wijesoma2004road} to isolate points of depth discontinuity or break points within each such ring. As illustrated in Figure~\ref{fig:breakpoint}, for a given triplet of consecutive points, $P = \{p_{i}, p_{i+1}, p_{i+2}\}$ having distances $D = \{d_{i}, d_{i+1}, d_{i+2}\}$ from the LiDAR origin, we utilise the LiDAR's horizontal angular resolution angle $\theta$, and the measured distances $d_{i}, d_{i+1}$ at $p_{i}$ and $p_{i+1}$ respectively to predict the distance $\mathit{d_p}$ at $p_{i+2}$. 
\begin{equation}
    \mathit{d_p} = \frac{d_{i}d_{i+1}}{2d_{i}\cos{\theta} - d_{i+1}}
\end{equation}
The point $p_{i+2}$ is categorised as a break point if the difference between the predicted distance $\mathit{d_p}$ and measured distance $d_{i+2}$ is beyond a certain threshold $\mathbf{d_{th}}$.
\begin{equation}
    |d_{i+2} - \mathit{d_{p}}| \geq d_{th}
\end{equation}

We then filter out the breakpoints detected outside the road. For sequences where road curbs and pedestrian pathways are elevated from the road plane, we use our detected breakpoints to isolate the road segments, ring wise within the Point Cloud, in a similar fashion to what has been discussed in \cite{wijesoma2004road}. We use a pre-trained SqueezeSeg~\cite{wu2018squeezeseg} for road segmentation when the curbs are not available. Our method is agnostic to the choice of road segmentation algorithm, and any pre-trained algorithm either in image or LiDAR space can be used for the task. 



\begin{figure}[t]
\includegraphics[width=0.5\textwidth]{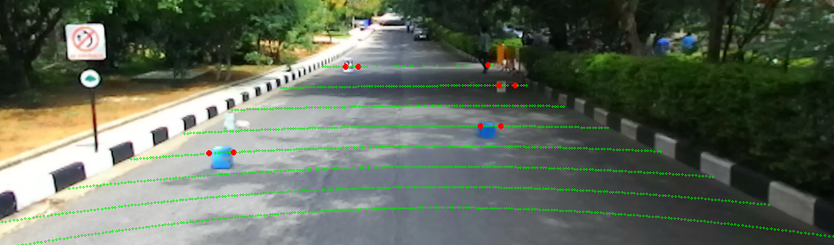}
\caption{LiDAR points projected onto the image space. Points within each pair of breakpoints (denoted in Red) represent a small obstacle segment (AB).}
\label{fig:vis_breakpt}
\end{figure}

\begin{figure*}
\centering
\includegraphics[width=0.97\textwidth]{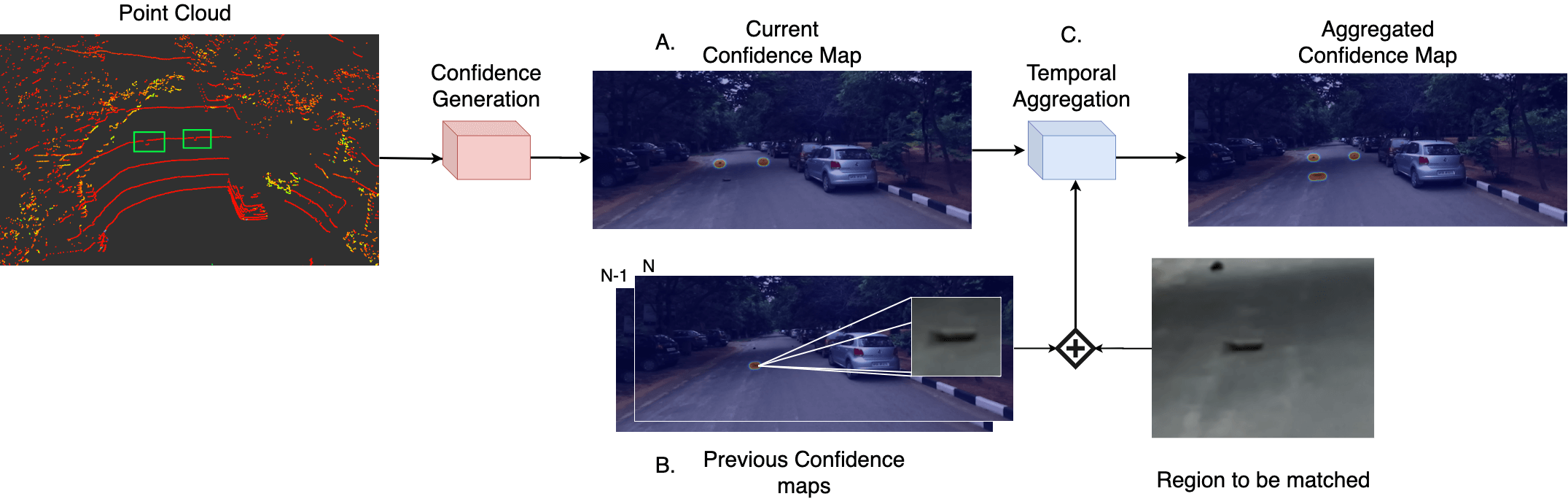}
\caption{Temporal Aggregation. (A) Current Confidence Map generated by break point detection and segment isolation. Note that the two discontinuities in the Point Cloud (depicted within green rectangles) correspond to current Confidence Map. A third obstacle is missed by the current LiDAR scan. (B) A template of the third small-obstacle is retrieved from contexts given in the past frames. (C) Using the template extracted and the current image, temporal aggregation is performed to generate an aggregated temporal confidence mask}
\label{fig:flow_temporal}
\end{figure*}

\textbf{Small Obstacle Segment Isolation:}
We define gradient direction at a breakpoint A as $G(A) = sign(d_{i+2} - \mathit{d_{p}})$. A consecutive pair of breakpoints A and B are said to belong to an obstacle segment if $G(A)$ is negative and $G(B)$ is positive. Small obstacle segments are then filtered out based on the horizontal/azimuthal angular resolution between the breakpoints A and B (example detected small obstacle segments as a pair of breakpoints are illustrated in Figure~\ref{fig:vis_breakpt}). This spread/resolution differentiates big obstacles such as cars from their smaller counterparts. For this dataset, we define this threshold as $\mathbf{2^\circ}$ and set the value of $\mathbf{d_{th}}$ as \textbf{0.4}. This allows us to detect an obstacle as small as 15 cms in height from about 50 meters away.


\textbf{Confidence Maps:}
To generate a pixel level confidence map, we project the detected small obstacle segments onto the image plane. The projection is performed using the mapping $\phi: \mathbb{R}^{3} \rightarrow \mathbb{R}^{2}$: 

\begin{equation}
    \mathbf{x} = \phi(\mathit{X},\mathit{K},\mathit{T}).
\end{equation}

where $\mathit{X}$ is the set of 3-D LIDAR points belonging to a small obstacle segment and $\phi$ is the projection operator. $\phi$ is parametrized by camera intrinsic matrix ($\mathit{K}$) and extrinsics matrix ($\mathit{T}$). The set of projected points, $\mathit{x}$, then serve as anchors of our confidence regions, where each point in this set is assigned a confidence value of \textbf{1} and the confidence values in the neighbourhood of this point follow a \emph{Gaussian} distribution (with mean centered on the point and variance parameterized by $\sigma$).


\subsection{Temporal Propagation of Confidences}
\label{sec:temporal}

We further augment the confidence maps using the temporal propagation of the previously detected small obstacles. This step helps to gather context for obstacles which are missed by the sparse LiDAR in the current frame. The updated confidence map is then used as the input to the segmentation network. The propagation is performed in a two step process: (a) we select a Region of Interest (ROI) in image space around the projections of the previously LiDAR detected small obstacle segments and (b) we find the corresponding ROI in the current frame using template matching in RGB space. We reduce the search space during template matching using the coarse odometry estimates between the past and the current frame. An example is illustrated in Figure~\ref{fig:flow_temporal}. 

\section{SMALL OBSTACLE DATASET}
\label{sec:Data}

We introduce the Small Obstacle Dataset to the community considering the importance of this problem for on-road navigation. Occurrences of small obstacles like bricks, granite rocks, stones may not be too uncommon in some parts of the world. Such objects are often on roads adjacent to or bordering large scale construction sites. Sometimes it is possible that sentient beings such as cats and dogs may be asleep on the roads. To this end, we curate a novel IIIT Small Obstacle Dataset encompassing significant variations in terms of obstacles, road types and lighting. 


\subsection{Description}
The dataset includes monocular RGB images and synchronized LiDAR scans with Odometry information. It comprises of 15 sequences in total and utilises a varied set of small obstacle instances. The details in terms of total images and train/val/test splits are given in Table~\ref{tab:dataset} We used different roads and different set of obstacles while recording the train, val and test sequences. Test split is kept to be most challenging in terms of turns, occlusions and shadows to better evaluate the generalizability. We kept the definition of a small obstacle as an object whose longest dimension is less than the diameter of a standard car wheel - which is around 21 inches.\\

\subsubsection{\textbf{Sensor and Software Setup}}
The Sensor and Software setup for recording the data is as follows
\begin{itemize}
    \item ZED Stereo camera (only left feed).
    \item Velodyne Puck (VLP-16).
    \item Vehicle - Mahindra E2O electric car.
    \item ROS Kinetic
\end{itemize}

\begin{table}[h!]
\centering
\begin{tabular}{|l|c|c|c|c|}
\hline
\textbf{Split} & \textbf{\# seq.} & \textbf{\# Images} & \textbf{with curbs} & \textbf{without curbs} \\

\hline\hline
     \textit{Train} & 9 & 1937 & 6 & 3\\
     \textit{Validation} & 4 & 530 & 2 & 2\\ 
     \textit{Test} & 2 & 460 & 1 & 1\\
     \hline\hline
     \textbf{Total} & $\mathbf{15}$ & $\mathbf{2927}$ & 9 & 6\\
\hline
\end{tabular}
\caption{Description of the Small Obstacle Dataset}
\label{tab:dataset}
\end{table}

\subsubsection{\textbf{Annotations}}
\label{Annotate}
We provide accurate pixel-level semantic annotations specific to the task of small obstacle segmentation for a selected set of images for all the sequences (sampling more densely around small obstacles). The total number of annotated images is 2927(Table~\ref{tab:dataset}). We also release all the raw sequences. The dataset is segregated into three classes, namely, Road, Off-Road and Small-Obstacle. Everything except road and small obstacles is labelled as off-road class (buildings, cars, pedestrians etc.). The Small-Obstacle class is further annotated with nine different types of sub-classes (stones, bricks, plastic, dogs etc.). 

\subsection{Improving LiDAR-Camera Calibration}\label{Haus}
LiDAR and Camera provide complementary streams of information in depth and colour and have become an indispensable part of the perception module for autonomous driving. To make vital environmental inferences, it is essential that they must be well calibrated so that their mutual information can be fused together. The problem hence involves finding the 6-DoF rigid body transformation matrix which allows for flexible transformation between their respective co-ordinate frames. This is done by solving for point-correspondences either between 2D-3D points i.e pixel co-ordinates in Camera frame and 3-D points in Point-Cloud or through 3D-3D associations. In the latter, specific setups like AR based ArUCo markers are used, which facilitate retrieving the 3-D coordinates in the camera frame through encoded patterns. We use this method for our setup utilising ArUco markers as detailed in \cite{dhall2017lidar}.

\begin{figure}[t]
\centering
\includegraphics[width=0.4\textwidth]{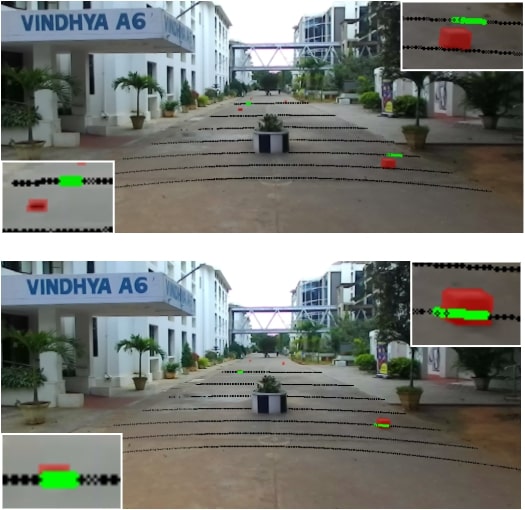}
\caption{(a) Point Cloud projection using calibration matrix obtained through \cite{dhall2017lidar}. (b) Improved calibration using our method. We can observe how projection error becomes more significant for farther obstacles.}
\label{fig:hausdorff_vis}
\end{figure}

While the method is specifically adapted to work with sparse 16 channel LiDAR over other methods - which use a denser 64 channel LIDAR(for eg. used in KITTI dataset~\cite{Kitti}), it still has its limitations. It relies on recognition of the patterns in ArUco marker to calculate the 3-D correspondences for the Camera frame, therefore the markers are kept within a few metres of the setup during the optimisation process. We observed that the calibration matrix obtained using point correspondences, estimated using this method(i.e corners of the marker), were not accurate at greater distances($\>$50m), posing a significant problem for our task (an example is illustrated in Figure~\ref{fig:hausdorff_vis}). We therefore propose a method to do fine refinement over the extrinsics parameters specifically for tasks where precise calibration is essential for sensor fusion. Below we explain our methodology and show Quantitative results in~\ref{tab:6} of improvement over the original task with this refined calibration method.

\textbf{2D-3D Point Correspondence:} The main intuition of our approach is that if the calibration is accurate, the small obstacle segments detected in 3D LiDAR point cloud when projected onto 2D image space should intersect with ground truth annotations of the small obstacle class. Formally, let $\mathit{S_1}$ denote the set of 3D points falling on multiple small obstacles and $\mathit{P_2}$ denote the set of 2D points belonging to the corresponding annotated pixel label in Image. We optimise over the extrinsic calibration parameters such that the projection set $\mathit{P_1 \in \mathbb{R}^2}$ of $\mathit{S_1}$ obtained through (\ref{eqn:4}), has the following property: $\mathbf{P_1} \subseteq \mathbf{P_2}$, i.e. all detected small obstacle segments in the point cloud lie within their respective pixel label bounds. The transformation operation $\phi: \mathbb{R}^{3} \rightarrow \mathbb{R}^{2}$ is defined as:
\begin{equation}\label{eqn:4}
    \begin{pmatrix} x\\ y \end{pmatrix} = \pi\begin{pmatrix} R\begin{pmatrix} X\\ Y\\ Z \end{pmatrix} + t \end{pmatrix}
\end{equation}
where $\pi$ subsumes the camera intrinsics ($f_x$,$f_y$,$c_x$,$c_y$) and R, t are the rotation and translation matrix in SO(3) respectively.

Let the 1 x 6 vector $\xi$ = $(\nu,\omega) \in$ se(3) denote our initial coarse calibration extrinsics where [$\nu_x,\nu_y,\nu_z$] is the translation vector and [$\omega_x,\omega_y,\omega_z$] is the rotational vector along their respective axis. The rotational vector $\omega$ $\in$ so(3) can be converted to R $\in$ SO(3) using the exponential map exp: so(3) $\to$ SO(3); $\omega$ $\to$ $e^\omega$.

\begin{equation}\label{eqn:5}
    R = I + \frac{w}{||w||}\sin{||w||} + \frac{w^2}{||w||^2}(1 - \cos{||w||^2})
\end{equation}

 According to our chosen point correspondences stated above, we now define the projection loss through Hausdorff distance.
 
\textbf{Hausdorff Projection Loss:} The directed Hausdorff distance between two sets $\mathbf{P_1}$ and $\mathbf{P_2}$ is defined as:
\begin{equation}\label{eqn:6}
    d(P_1, P_2) = \underset{x \in P_1}{\sup} \underset{y \in P_2}{\inf} \vert\vert x - y \vert\vert
\end{equation}

The loss function thus gives a measure of the distance the set $P_1$ has to travel in Image space so as to be completely contained inside of $P_2$. Gradients with respect to each of the 6 elements of the se(3) transformation vector $\xi$ can then be back-propagated while minimising the loss with any gradient based optimizer. For the experiment, we sample a few frames from each sequence of the Training split of the Dataset and use Adam~\cite{kingma2014adam} optimizer with a learning rate of $1\text{e-}5$.

\section{EXPERIMENTS AND DISCUSSION}
\label{sec:EnD}

\subsection{Evaluation metrics}
We evaluate our model's performance on both instance-level and pixel-wise metrics: 

\textbf{Instance Detection Rate (IDR):} Instance-level detection rate (IDR) is defined as the fraction of obstacle instances, taken across the dataset, which are detected by the network. For this metric, an instance is marked correctly detected if more than 20\% of the pixels of the predicted obstacle overlaps with the ground truth of that instance. For extracting instances from pixel-level predictions we use a similar approach to~\cite{mergenet}. The IDR metric is formally calculated as:
\begin{equation}
    \mathit{IDR} = \frac{TP_{obstacle}}{TI_{obstacle}}
\end{equation}
Where a true positive, $TP_{obstacle}$, represents a set of predicted instances that have an overlap greater than a threshold of 0.2 with the ground-truth. $TI_{obstacle}$ is the  total instances of the obstacle class, taken across the entire dataset.

\textbf{instance False Detection Rate (iFDR)}
This metric is defined as the fraction of instances that have been incorrectly detected as small obstacles:
\begin{equation}
    \mathit{iFDR} = \frac{FP_{obstacle}}{PRED_{obstacle}}
\end{equation}
Where a $FP_{obstacle}$ is defined as any predicted small obstacle prediction instance that has no overlap with ground truth and $PRED_{obstacle}$ are the total number of instances predicted as small obstacles.

\textbf{Pixel Detection Rate (PDR)}
$\mathit{PDR}$ is the fraction of small obstacle pixels that have been correctly predicted.
\begin{equation}
    \mathit{PDR_{obstacle}} = \frac{TPX_{obstacle}}{GTX_{obstacle}}
\end{equation}
where $TPX_{obstacle}$ denotes the total number of correctly predicted small obstacle pixels and $GTX_{obstacle}$ is the total number of small obstacle pixels in the image.

\textbf{mean Intersection over Union (mIoU)}
The mean Intersection over Union (mIoU) is a commonly used metric for semantic segmentation. It measures the average Intersection over Union across all classes.
\begin{equation}
    \mathit{IoU = \frac{TPX}{TPX+FPX+FNX}}
\end{equation}
Where, $TPX$, $FPX$ and $FNX$ are pixel level True Positives, False Positives and False Negatives for the small obstacle class. 

\begin{table}
\centering
\begin{tabular}{|l|l|c|c|c|c|}
\hline
\multirow{2}{*}{\textbf{Network}} & \multirow{2}{*}{\textbf{Method}} & \multicolumn{2}{|c|}{\textbf{Instance-level}} & \multicolumn{2}{|c|}{\textbf{Pixel-level}}\\

& & IDR & iFDR & PDR & mIoU \\

\hline\hline
     \multirow{3}{5em}{\textbf{DeepLab-V3+}\cite{DeepLab}} & \textit{Image} &  0.39 & 0.28 & 0.37 & 0.73\\
     & \textit{Image + CM} & 0.50 & 0.26 & 0.45 & 0.74\\ 
     & \textit{Image + CM + TP} & \textbf{0.60} & \textbf{0.18} & \textbf{0.60} & \textbf{0.76}\\
     
\hline
    \multirow{3}{5em}{\textbf{HRNet}\cite{HRNet}} & \textit{Image} &  0.44 & 0.25 & 0.27 & 0.70\\
     & \textit{Image + CM} & 0.47 & 0.25 & 0.32 & 0.70\\
     & \textit{Image + CM + TP} & \textbf{0.63} & \textbf{0.21} & \textbf{0.51} & \textbf{0.73}\\ 
\hline
\end{tabular}
\caption{Performance comparison between various inputs to the semantic segmentation architectures.}
\label{tab:results}
\end{table}

\subsection{Experiments}
\label{sec:Expt}
We compare our method with two monocular semantic segmentation baselines. We present thorough ablations to motivate each component of the proposed pipeline. The evaluations are performed on the test set of the proposed IIIT Small Obstacle Dataset. Each experiment is trained on 2 Nvidia GTX-1080ti GPUs for 15 epochs. A batch size of 6 is kept during training.
 
As to the choice of the segmentation network, we picked two state of the art networks DeepLabV3+~\cite{DeepLab} and HRNet~\cite{HRNet} on CityScapes benchmark. They also made a suitable choice due to availability of pre-trained model/weights and well documented reproducible code base. We perform three set of experiments considering pre-trained DeepLabV3+ and HRNet as base networks. The pre-training is performed on the CityScapes dataset. 

 \textbf{Image baseline: } Our baseline for comparison is the performance of base network on the Small Obstacle Dataset, when trained on images only. We fine-tune a pre-trained base network on the proposed dataset to give a 3 class per pixel prediction. We supervise the network using the Cross-Entropy loss with inverse frequency rebalancing. 
 
\textbf{Image + LiDAR confidence masks:} Next, we modify the base-network by initialising a new channel for the input layer so as to accept a 4D input tensor. We then train the network on images concatenated with confidence masks (generated using LiDAR projections) along the channel dimension. This experiment only considers the current LiDAR for generating the confidence maps.

 \textbf{Image + LiDAR confidence masks + TP:} This experiment is similar to (Image + LiDAR) in network configuration. However, the temporal context (4 previous frames) is considered to generate the confidence maps (to compensate of obstacles missed in the current frame). 

\subsection{Results}
In Table~\ref{tab:results} we evaluate the performance of our framework. All metrics are reported for the test set. The DeepLab-V3+ image-only baseline achieves an $IDR$ of 0.39. Our proposed method of temporally aggregated confidence maps shows a \textbf{53\%} improvement to an $IDR$ of 0.60, while reducing the $iFDR$ by \textbf{35\%} to a score of 0.18. Furthermore, this improvement comes with a \textbf{4\%} increase to 0.76 in the mIoU metric.

A similar trend is seen when our framework is trained with the HRNet segmentation network. The HRNet image-only baseline achieves a higher $IDR$ than DeepLab-V3+ at 0.44. When trained with our proposed temporally aggregated confidence masks along with images, performance improves by \textbf{43\%} to a score of 0.63, while the $iFDR$ drops by \textbf{16\%} to a score of 0.21.

We further note the increase in the $IDR$ metric between utilising obstacle confidences only from the current frame (\textit{Image + CM}), and temporally propagating such confidences from previous frames ({Image + CM + TP}). As can be seen in Table~\ref{tab:results}, there is a \textbf{20\%} corresponding increase from 0.50 to 0.60 by augmenting the confidence map using temporal observations. This reinforces the importance of temporal confidence propagation. 

\begin{figure*}[t]
   \centering
\begin{tabular}{cccc}
\includegraphics[width=0.20\textwidth]{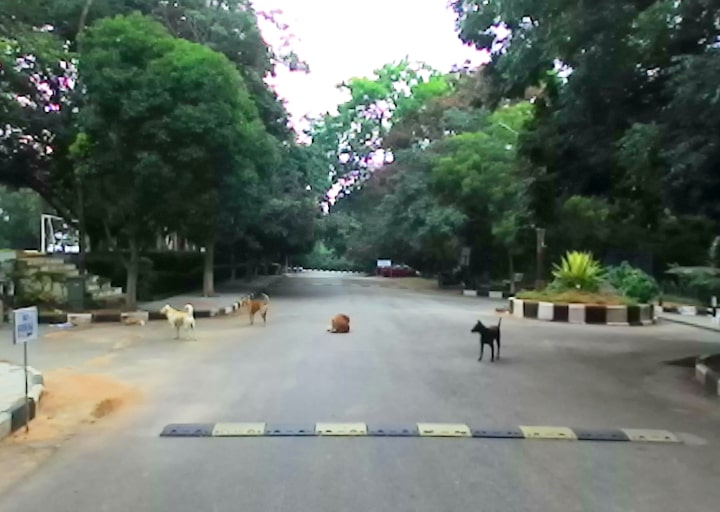}&
\includegraphics[width=0.20\textwidth]{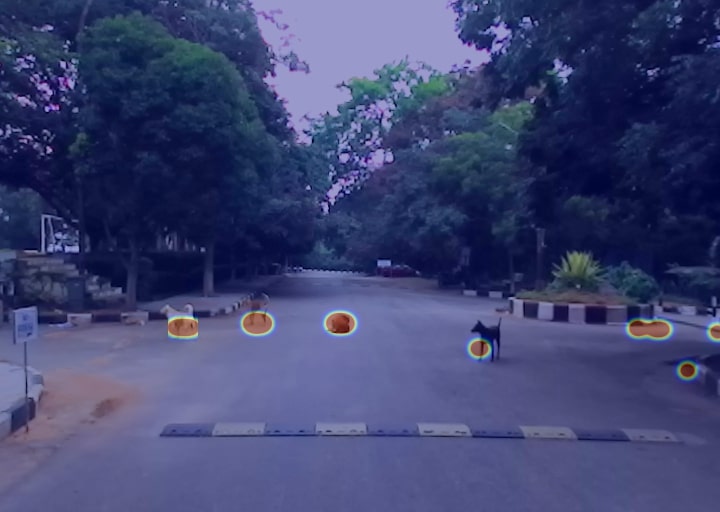}&\includegraphics[width=0.20\textwidth]{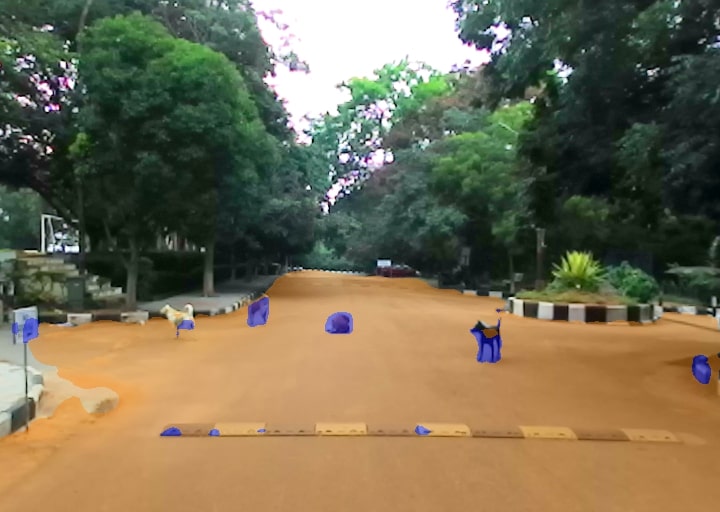}&\includegraphics[width=0.20\textwidth]{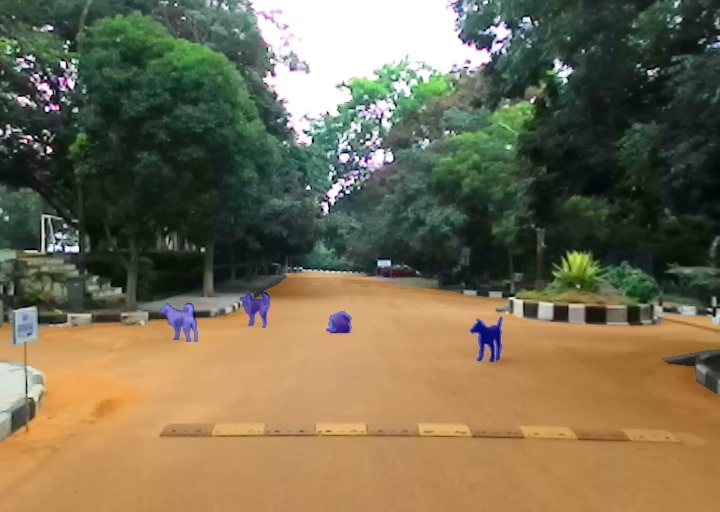}\\
\includegraphics[width=0.20\textwidth]{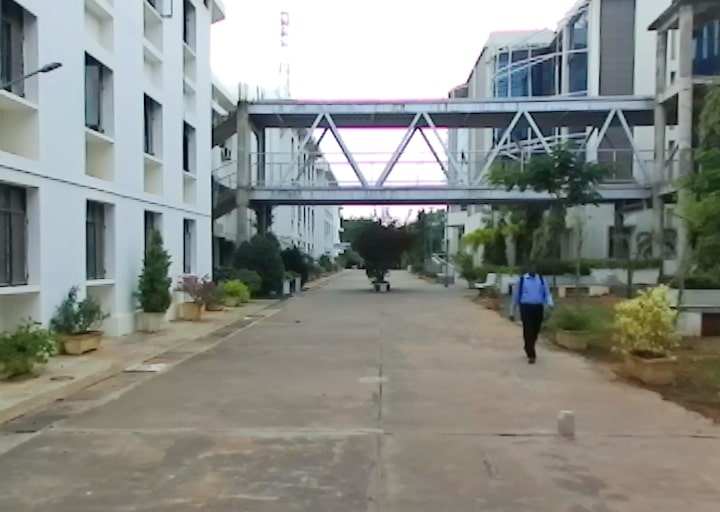}&
\includegraphics[width=0.20\textwidth]{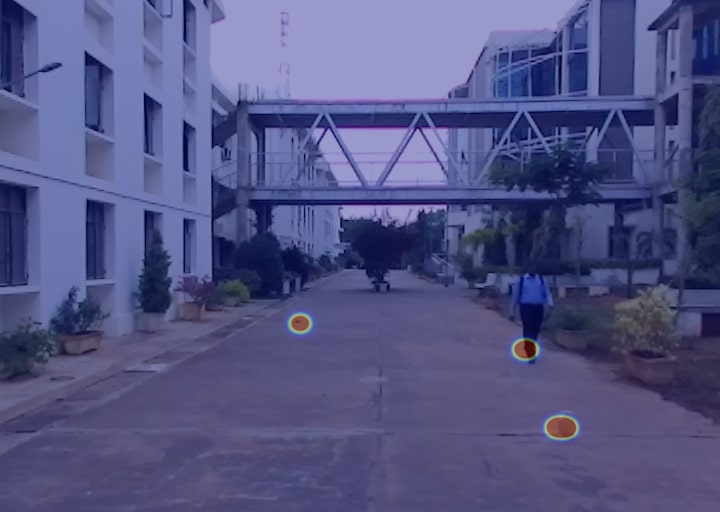}&\includegraphics[width=0.20\textwidth]{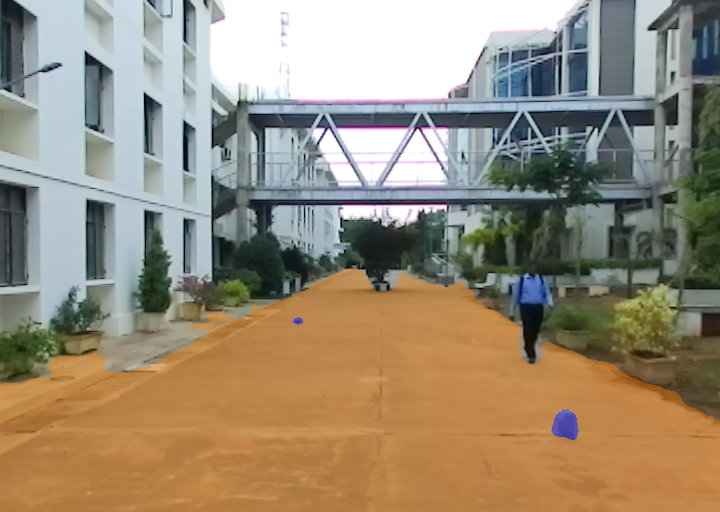}&\includegraphics[width=0.20\textwidth]{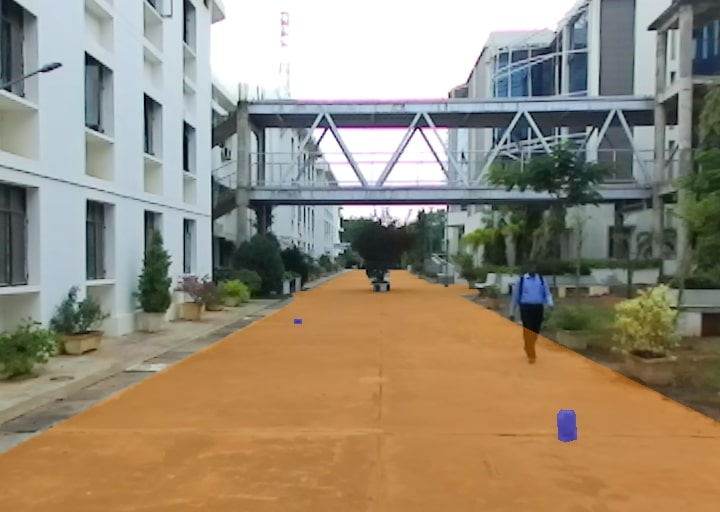}\\
\includegraphics[width=0.20\textwidth]{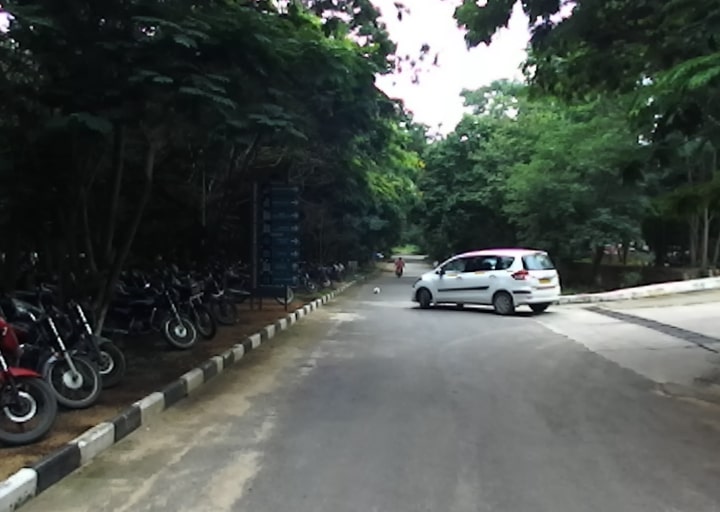}&
\includegraphics[width=0.20\textwidth]{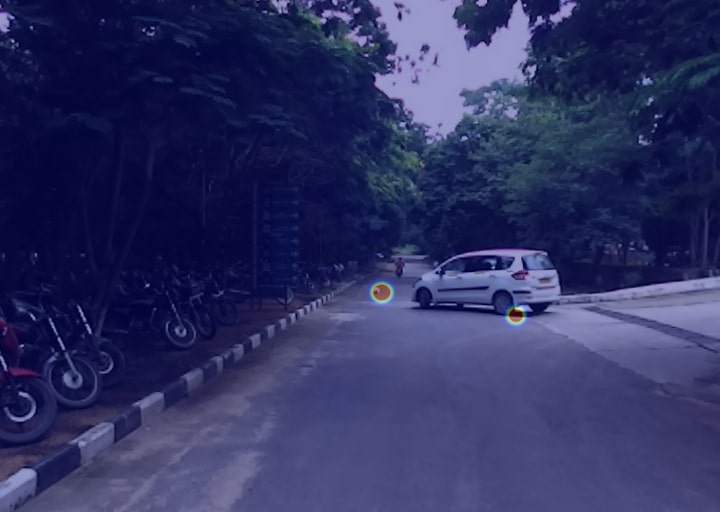}&\includegraphics[width=0.20\textwidth]{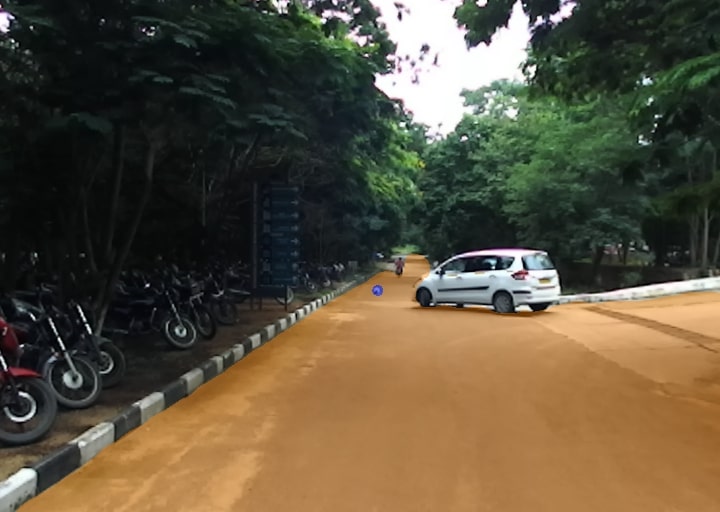}&\includegraphics[width=0.20\textwidth]{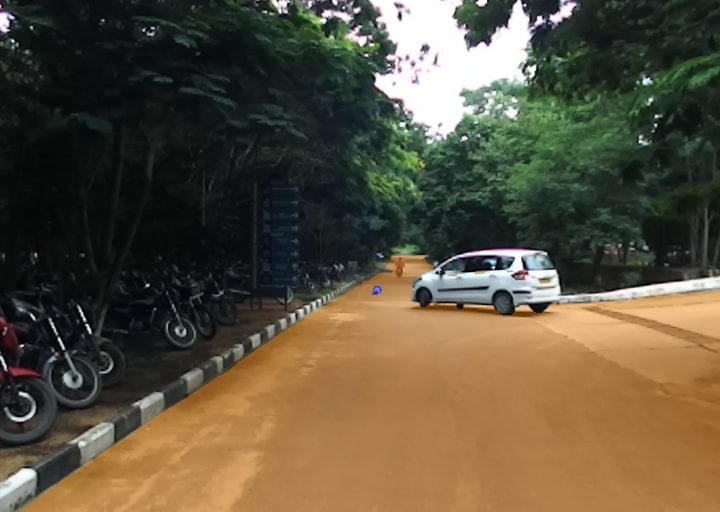}\\
\includegraphics[width=0.20\textwidth]{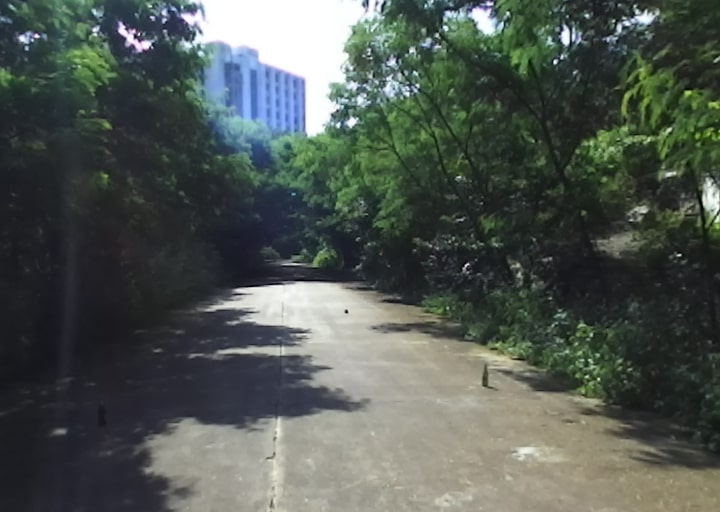}&
\includegraphics[width=0.20\textwidth]{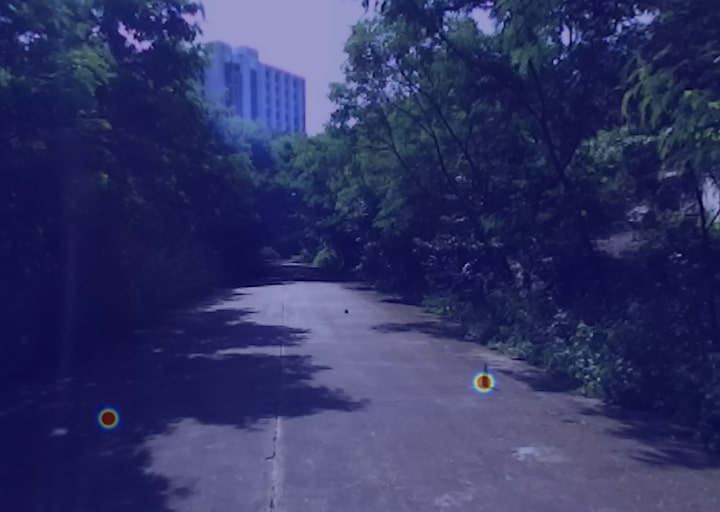}&\includegraphics[width=0.20\textwidth]{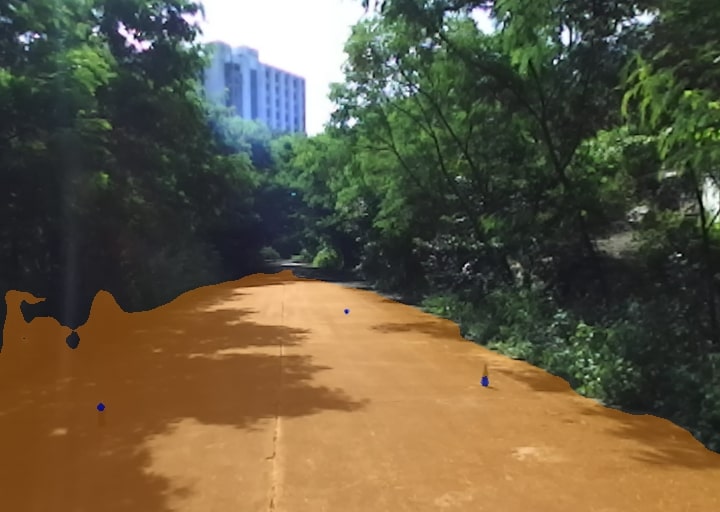}&\includegraphics[width=0.20\textwidth]{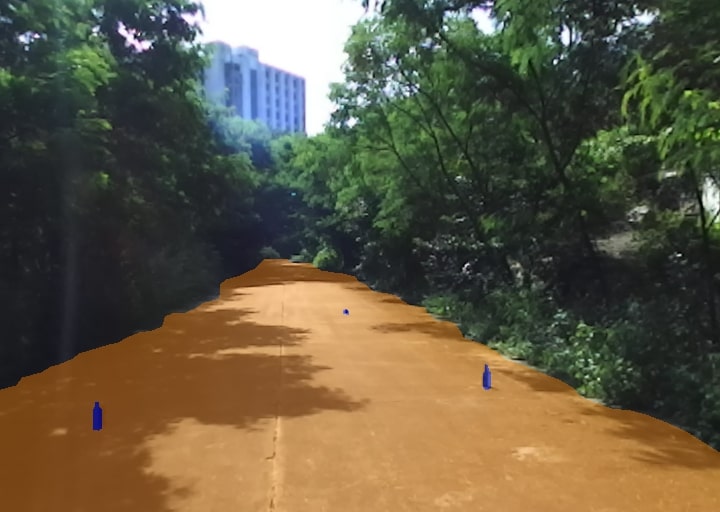}\\
{\small (a)} & {\small (b)} & {\small (c)} & {\small (d)}\\
\end{tabular}
    \caption{(a) Image. (b) Confidence Map (c) Prediction. (d) Ground-truth}
\label{fig:dataset_vis} 
\end{figure*}

We present a few qualitative results across our test and validation splits in Fig. \ref{fig:dataset_vis}. It can be observed that even when there is a high confidence value on legitimate regions within the image (in Fig. \ref{fig:dataset_vis}, legs of a man, wheel of a car), the joint representation has learnt to classify these regions as non-obstacle(road/off-road).


     


\begin{table}
\centering
\begin{tabular}{|l|c|c|}
\hline
\textbf{Method} & IDR & PDR\\
\hline\hline
     \textit{RANSAC} &  0.424 & 0.43\\
     \textit{Conv-1D} & 0.421 & 0.373\\ 
     \textit{Geometric (ours)} & 0.50 & 0.45\\
\hline
\end{tabular}
\caption{A comparison of various small obstacle detection techniques on Point Cloud.}
\label{tab:4}
\end{table}
\quad

\section{ABLATION STUDY}
We show the following ablation experiments across our three main contributions: (a) Detection of small obstacles in the  LiDAR Point Cloud, (b) Temporal propagation of confidences and (c) improved extrinsics calibration. All of the results presented below use DeepLab-V3 as the base network architecture.

\subsubsection{Detection of Obstacle Segments}
We experiment with multiple approaches for detecting a small obstacle within a Point Cloud ring. In Table-\ref{tab:4} we show our results with three such methods for the Image + Confidence Map Baseline. We first experiment with a plane-fitting method where points falling on small obstacles are classified as outliers through Random sample consensus \cite{fischler1981random}. Through this we can successfully detect bigger small obstacle instances however it fails on small instances such as rocks,bricks etc. We then study a learning based method where a small 1-D convolutional network is trained to classify each point within the curb boundaries as a road or an obstacle point. 
The best results were obtained using geometric approach (Sec~\ref{sec:geometric}) and was thus used in our final pipeline. 


\subsubsection{Temporal Propagation of Confidences}
We tried two different methods for temporal propagation and observed their impact on the final output (Table~\ref{tab:5}). In Forward Projection, we directly project obstacle detections in previous 4 frames onto the current Image frame using Odometry information calculated through LiDAR Odometry and Mapping\cite{zhang2014loam}. However due to less than desired accuracy in the Odometry estimate this doesn't localize the confidence map to the true small obstacle instance on the Image and shows minor improvement. The proposed temporal aggregation (Sec~\ref{sec:temporal}) gives much improved results.


\subsubsection{Improved Extrinsics Calibration}
Here we study the effect of better calibration on our method comparing the results on Image + Confidence Map baseline. In Table-\ref{tab:6} it can be seen how better calibration significantly affects the detection of small obstacle instances. It is to be noted that increasing the spatial spread of the Confidence region/map with coarse extrinsics (Pre-Hausdorff) does not translate to a comparable performance with a better set of extrinsics obtained using our method. This highlights the need for calibration refinement in our pipeline.

\begin{table}
\centering
\begin{tabular}{|l|c|c|c|}
\hline
\textbf{Method} & IDR & iFDR & PDR\\
\hline\hline
     \textit{Forward Projection} &  0.52 & 0.23 & 0.43\\
     \textit{Temporal Aggregation} & 0.60 & 0.18 & 0.60\\ 
\hline
\end{tabular}
\caption{A comparison of methods for temporal propagation of detected obstacles.}
\label{tab:5}
\end{table}

\begin{table}
\centering
\begin{tabular}{|l|l|c|c|c|}
\hline
\textbf{Extrinsics Parameters} & \textbf{$\sigma$} & IDR & iFDR & PDR\\
\hline\hline
     \multirow{2}{6em}{\textbf{Pre-Hausdorff}} & 5 &  0.412 & 0.263 & 0.363\\
     & 7 & 0.426 & 0.217 & 0.36 \\
     \hline
     \multirow{2}{6em}{\textbf{Post-Hausdorff}} & 5 &  0.50 & 0.26 & 0.45\\
     & 7 & 0.49 & 0.22 & 0.464 \\
\hline
\end{tabular}
\caption{Comparison on the detection performance between Coarse and Fine calibration obtained using our method. $\sigma$ denotes the variance of the Gaussian blur.}
\label{tab:6}
\end{table}

\section{CONCLUSIONS}
Our work focuses on the problem of detecting small obstacles lying on road by fusing monocular RGB and LiDAR data. Our experiments show that IDR increases by 40-50$\%$ when a given image only segmentation network is augmented with Confidence Maps provided by LiDAR. Thorough experiments using two state of the art base networks DeepLABV3+ and HRNet demonstrate the efficacy of our approach. We also show that small obstacle ground truth annotations can be exploited to improve the extrinsic calibration, which in turn improves the IDR (Table-~\ref{tab:6}). We further present thorough ablation studies to justify the design choices. Overall, our method is able to detect 73\% of the small obstacles (as small as 15cm high) within the range of 50 meters. 


\bibliographystyle{IEEEtran}
\bibliography{references}
\end{document}